\documentclass{article}

\usepackage{amsmath}
\usepackage{mdframed}
\usepackage{caption}
\usepackage{authblk}
\usepackage{graphicx}
\usepackage{float}

\title{Tensor-Based Backpropagation in Neural Networks with Non-Sequential Input}
\date{}
\author{\small Hirsh R. Agarwal, Andrew Huang}
\affil{\footnotesize \emph{University of Edinburgh, AndPlus LLC}}

\begin{document}
	\pagenumbering{gobble}
	\maketitle
	\pagenumbering{arabic}
\begin{abstract}
Neural networks have been able to achieve groundbreaking accuracy at tasks conventionally considered only doable by humans. Using stochastic gradient descent, optimization in many dimensions is made possible$^{[1]}$, albeit at a relatively high computational cost. By splitting training data into batches, networks can be distributed and trained vastly more efficiently and with minimal accuracy loss. We have explored the mathematics behind efficiently implementing tensor-based batch backpropagation algorithms. A common approach to batch training is iterating over batch items individually. Explicitly using tensor operations to backpropagate allows training to be performed non-linearly, increasing computational efficiency.
\end{abstract}

\section{Introduction}
\begin{figure}[h]
\centering
\caption{Example Neural Network Structure}
\label{fig:network}
\includegraphics[scale=.45]{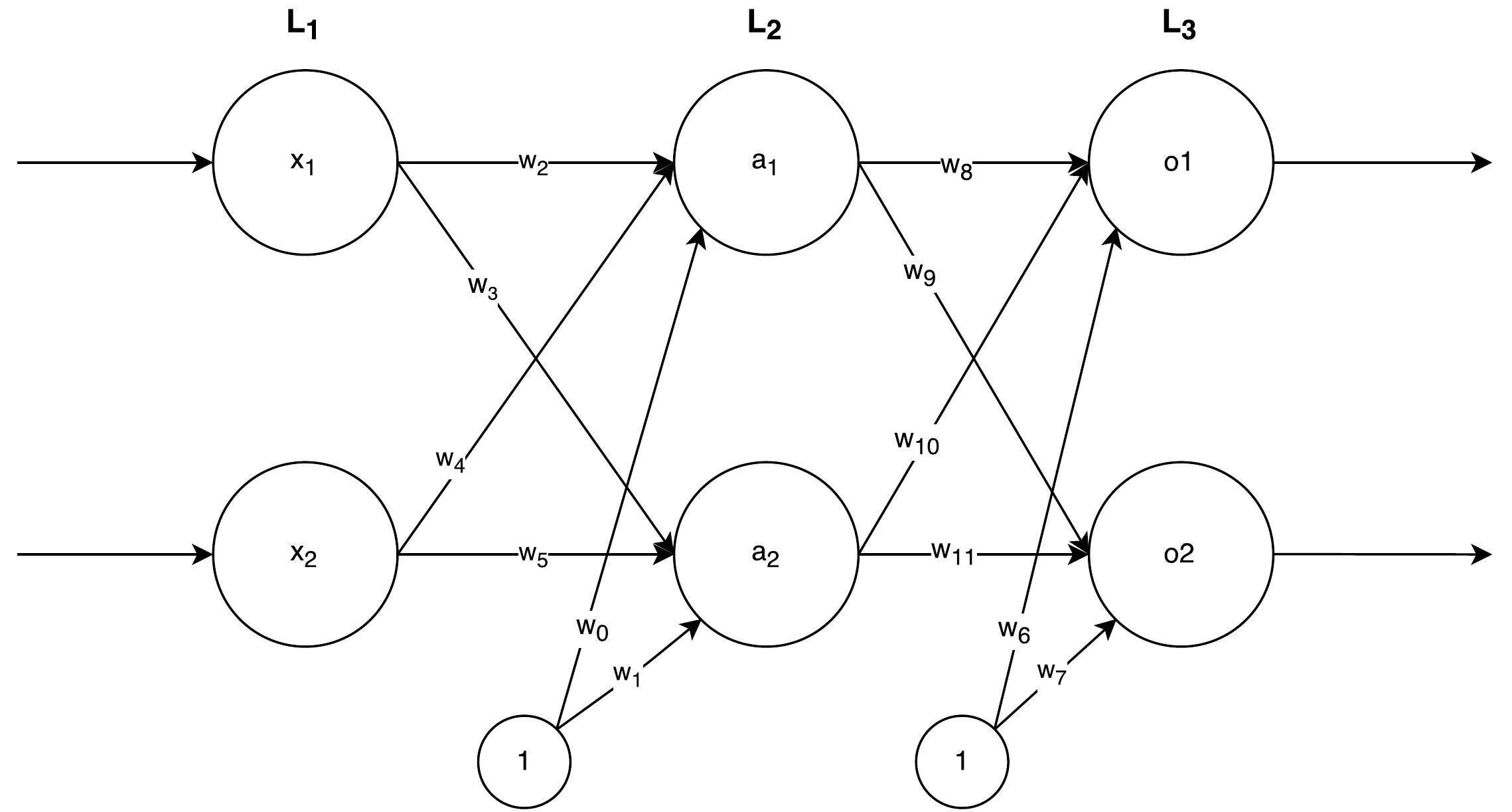}
\end{figure}

For a sequential data input neural network, we start with an example of forward propagation through a sequential input model. The input is a matrix containing two values
$[\begin{matrix}
x_1 & x_2
\end{matrix}]$. 
Our neural network is constructed with one hidden layer ($L_2$) and an output layer ($L_3$), each with two neurons (Figure \ref{fig:network}).
\\
\\
Given the structure of the network, the first layer ($L_1$) has a 2x3 weight matrix. In order to feed the input layer through this hidden layer, we take the dot product of the two matrices. The first values of the weight array ($w_0, w_1$) correspond to the bias values. We include 1 as the first input value in order to represent the bias input. Including a bias in the input matrix prevents the need to compute an error value separately, because it is treated as an extension of the previous layer. This bias is added in the same way for every layer in the network. The values $a_1$ and $a_2$ both have a differentiable activation function ($\sigma$) applied to them before being passed onto the next layer. Here, a sigmoid activation function is used: ($\sigma(x) = \frac{1}{1+e^-x}$).
\\
\[
[\begin{matrix}
1 & x_1 & x_2
\end{matrix}]
\cdot
\left[
\begin{matrix}
w_0 & w_1\\
w_2 & w_3\\
w_4 & w_5
\end{matrix}
\right]
=
[\begin{matrix}
a_1 & a_2
\end{matrix}]
\]
\\
\\
\\
After the input has been passed through the first hidden layer, the activation function is applied to all of the output values ($a_n := \sigma(a_n)$). The output from the activation function with another added bias (1) is now passed through the next set of weights. Again, the top weight values \((w_6, w_7)\) represent the weights on the biases.
\\
\[
[\begin{matrix}
1 & a_1 & a_2
\end{matrix}]
\cdot
\left[
\begin{matrix}
w_6 & w_7\\
w_8 & w_9\\
w_{10} & w_{11}
\end{matrix}
\right]
=
[\begin{matrix}
o_1 & o_2
\end{matrix}]
\]
\\
This yields the two output values, $(o_1,\, o_2)$. After being passed through another activation function, these are the model outputs.
\\
\hrule
\hrule
\subsection{Sequential Input Error}
The output values, \(\sigma(o_1, o_2)\), are defined as matrix $\psi_1$ and our target values as vector $\hat{t} = [t_1\; t_2]$. As an example, we use the sum of squares error metric to define the model error. Depending on the objective, different error metrics might be more appropriate$^{[2]}$. Each output value will have an error denoted by matrix $\hat{E}$, with a scalar value E denoting the total error for the model.
\\
\[
\hat{E} = \frac{1}{2}[\hat{t}-\psi_1]^2 = 
\frac{1}{2} [\begin{matrix}
t_1-o_1& t_2-o_2
\end{matrix}]^2
\]
\[
E = \Sigma(\hat{E})
\]
\subsection{Batch Input}
The process for forward propagating batched input is much the same as sequential input. Because all of the operations are vectorized, each batch item can be distributed to an independent computational system. 
The input matrix contains two input pairs, one pair for each batch item. Similar to the sequential input example, an extra 1 is added to the start of each batch item to act as the bias value.
\\
\[
\left[
\begin{matrix}
1 & x_1 & x_2\\
1 & y_1 & y_2
\end{matrix}
\right]
\]
\\
The same network structure (displayed in Figure \ref{fig:network}) is used, meaning the weight matrix remains unchanged. As with the sequential input, taking the dot product of the input vector and the weight matrix moves the data forward through the network. The primary difference is that the output matrix has an extra dimension, representing an output for each item in the input batch.
\\
\[
\left[
\begin{matrix}
1 & x_1 & x_2\\
1 & y_1 & y_2
\end{matrix}
\right]
\cdot
\left[
\begin{matrix}
w_0 & w_1\\
w_2 & w_3\\
w_4 & w_5
\end{matrix}
\right]
=
\left[
\begin{matrix}
a_1 & a_2\\
b_1 & b_2\\
\end{matrix}
\right]
\]
\\
The output from the hidden layer has the activation function applied ($a_n := \sigma(a_n)$) and bias values added. This value can be passed in the same manner through the next layer to yield an output for each individual batch item.
\\
\[
\left[
\begin{matrix}
1 & a_1 & a_2\\
1 & b_1 & b_2
\end{matrix}
\right]
\cdot
\left[
\begin{matrix}
w_6 & w_7\\
w_8 & w_9\\
w_{10} & w_{11}
\end{matrix}
\right]
=
\left[
\begin{matrix}
o_{x1} & o_{x2}\\
o_{y1} & o_{y2}\\
\end{matrix}
\right]
\]

\subsection{Batch Input Error}
The process for computing the sum square error for a batch input is similar to computing the sum square error for sequential input. The output from the last layer ($L_3$) (with activation function $\sigma$), $\sigma(o_{x1}\; o_{x2}; o_{y1}\; o_{y2}$), will be stored as matrix $\psi_2$. Target matrix $t = [t_{x1}\; t_{x2}; t_{y1}\; t_{y2}]$ is a square matrix representing the target for each output value.\footnote{It is only square because our batch size and input size are the same. Generalized dimension is (batch size, input size).} The matrix for individual output error is defined as $\hat{E}$, similar to the sequential example. This differs from the sequential model in that the sum error ($\hat{E}$) is a vector with each value corresponding to one batch input.\footnote{n in these equations is equal to the number of inputs.}
\\
\[
\hat{E} = 
\frac{1}{2n}(t-\psi_2)^2 = 
\frac{1}{2n}
	\left[	
	\begin{matrix}
	t_{x1}-o_{x1} & t_{x2}-o_{x2}\\
	t_{y1}-o_{y1} & t_{y2}-o_{y2}
	\end{matrix}
	\right]^2
\]
\\
\\
Total sum error is calculated in the same way, however, rather than being the sum of the whole matrix, it must now be explicitly defined as the sum of each row. The per output error averaged over each of the inputs can also be computed from this matrix by taking the average of each column. 
This error matrix is represented slightly differently than for sequential input, given the extra dimension for each batch item.
\\
\[
E = \Sigma(\hat{E}_{ij}) = 
\left[
\begin{matrix}
E_{x1} & E_{x2}\\
E_{y1} & E_{y2}
\end{matrix}
\right]
\]
\section{Backpropagation}
\subsection{Overview}
With the total error from forward propagation, the adjustment values for each weight can be calculated. We use gradient descent to compute the weights for the output layer $(L_3)$. In order to change the weights for the earlier layers $(L_1,L_2)$, we use backpropagation, determining the amount the weights need to change based on how much they were contributing to the error. All of the operations are generalized to tensor operations, making the underlying math for sequential and batch training models the same. Computing the gradient of the error will yield an n-d matrix, in which each component of the gradient equates to an adjustment to the corresponding weight.
\\
\[
\nabla E =
\left[
\begin{matrix}
\frac{\partial E}{\partial W_{11}} & \frac{\partial E}{\partial W_{12}}   & .\,.\,. & \frac{\partial E}{\partial W_{1j}}\\
\frac{\partial E}{\partial W_{21}} & \frac{\partial E}{\partial W_{22}} & .\,.\,. & \frac{\partial E}{\partial W_{2j}}\\
\vdots & \vdots & \ddots & \vdots\\
\frac{\partial E}{\partial W_{i1}} & \frac{\partial E}{\partial W_{i2}} & .\,.\,. & \frac{\partial E}{\partial W_{ij}}
\end{matrix}
\right]
\]
\subsection{Weight Adjustments for Last Layer}
Computing the error change per weight is based on three factors: final output, input to the activation function, and layer inputs. Each factor steps backwards through the network until the weight that is being updated is encountered.
\\
\[
\nabla E =
\frac{\partial E}{\partial W} =
\frac{\partial Net}{\partial W} \cdot
( 
\frac{\partial E}{\partial \psi} \odot
\frac{\partial \psi}{\partial Net}
)
\]
\newenvironment{aside}
  {\begin{mdframed}[style=0,%
      leftline=false,rightline=false,leftmargin=2em,rightmargin=2em,%
          innerleftmargin=0pt,innerrightmargin=0pt,linewidth=0.5pt,%
      skipabove=7pt,skipbelow=7pt]\small}
  {\end{mdframed}}
\begin{aside}

\begin{center}
	\title{Aside.}
	\\
	In this paper, the $\odot$ symbol is used to describe the Hadamard product. The standard $\cdot$ between two matrices represents a dot product (inner product). The meaning of each symbol is made explicit in its context within the paper when used.
\end{center}
\end{aside}
The equation $\frac{\partial E}{\partial W}$ represents the change in error with respect to each weight in the weight matrix, i.e. how much a weight value is contributing to the error. This $\frac{\partial E}{\partial W}$ value describes how to update the weight matrix. $\frac{\partial E}{\partial \psi}$ shows the change in total error with respect to the output, with $\psi$ representing the final output after the activation function is applied. $\frac{\partial \psi}{\partial Net}$ is the change in output with respect to the net output, in this case, the input to the activation function.
\\
\\
Part of this equation is often simplified with the representation $\delta = \frac{\partial E}{\partial \psi} \odot \frac{\partial \psi}{\partial Net}$. This is known as the delta rule. The delta value ($\delta$) is useful to store as it is used in subsequent layers of backpropagation.

\subsection{Computing $\frac{\partial E}{\partial W}$} \label{procedure}
For the output layer, the computation of each partial derivative is dependent primarily on the output values from the network.
$\frac{\partial E}{\partial \psi}$ computes to the difference of the output and error values, represented as $\psi - \hat{E}$. The error matrix is used, rather than the total error, as it keeps information about the accuracy of each output; this allows each of them to be adjusted individually. These matrices are the exact same size, allowing for calculating element wise matrix difference. Using example values from the batch forward propagation section:
\\
\[
\hat{E} = 
\left[
	\begin{matrix}
	E_{x1} & E_{x2}\\
	E_{y1} & E_{y2}
	\end{matrix}
\right]
,
\;
\psi = 
\left[
	\begin{matrix}
	o_{x1} & o_{x2}\\
	o_{y1} & o_{y2}
	\end{matrix}
\right]
\]
\\
\[
\frac{\partial E}{\partial \psi} = 
(\psi - \hat{E}) = 
\left[
	\begin{matrix}
	o_{x1}-E_{x1} & o_{x2}-E_{x2}\\
	o_{y1}-E_{y1} & o_{y2}-E_{y2}
	\end{matrix}
\right]
\]
\\
The change created by the activation function, represented by $\frac{\partial \psi}{\partial Net}$, is equal to the derivative of the same function. As previously mentioned, for this example, a sigmoid activation function is used. Because $\psi = \sigma(Net)$, $\frac{\partial \psi}{\partial Net}$ is the derivative of the activation function.
\\
\[
\frac{d}{dNet}(\sigma) =
\frac{d}{dNet}
	\left(
		\frac{1}{1+e^{-Net}}
	\right)
=
\psi (1-\psi )
\]
\\
\[
\frac{\partial \psi}{\partial Net} =
\left[
	\begin{matrix}
	o_{x1}(1-o_{x1}) & o_{x2}(1-o_{x2})\\
	o_{y1}(1-o_{y1}) & o_{y2}(1-o_{y2})
	\end{matrix}
\right]
\]
\\
\\
With the values of $\frac{\partial E}{\partial \psi}$ and $\frac{\partial \psi}{\partial Net}$, the $\delta$ value can be calculated. Because the two matrices are of identical shape, the Hadamard product of the two matrices can be taken.
\[
\delta = 
\frac{\partial E}{\partial \psi}
\odot
\frac{\partial \psi}{\partial Net}
\]
For this layer ($L_3$), $\frac{\partial Net}{\partial W}$ can be calculated as the output weights from the previous layer ($L_2$) with a 1 included as the bias value. This matrix is exactly what is fed to the weights during forward propagation, except transposed.
\\
\[
\phi = 
\frac{\partial Net}{\partial W}
=
\left[
\begin{matrix}
1 & a_1 & a_2\\
1 & b_1 & b_2
\end{matrix}
\right]^T
=
\left[
\begin{matrix}
1 & 1\\
a_1 & b_1\\
a_2 & b_2
\end{matrix}
\right]
\]
\\
Now, all three values have been computed. The dimensions of $\left[\frac{\partial Net}{\partial W}\right]^T (\phi)$ and $\delta$ should match along their inner axes. For example, $\phi$ = (m, b) and $\delta$ = (b, n), where b will always be equal to the batch size.
In order to compute the weight change values, we take the inner tensor product  of the two matrices.
\\
\[
\frac{\partial E}{\partial W} = 
\phi \cdot \delta
=
\left[
\begin{matrix}
1 & 1\\
a_1 & b_1\\
a_2 & b_2
\end{matrix}
\right]
\cdot
\left[
	\begin{matrix}
	\delta_{x1} & \delta_{x2}\\
	\delta_{y1} & \delta_{y2}
	\end{matrix}
\right]
=
\left[
	\begin{matrix}
	\delta_{x1} + \delta_{y1} & \delta_{x2} + \delta_{y2} \\
	\delta_{x1} a_1 + \delta_{y1} b_1 & \delta_{x2} a_1 + \delta_{y2} b_1\\
	\delta_{x1} a_2 + \delta_{y1} b_2 & \delta_{x2} a_2 + \delta_{y2} b_2\\
	\end{matrix}
\right]
\]
\subsection{Updating Weights}
With the $\frac{\partial E}{\partial W}$ value computed, we can find $\Delta W$ value by multiplying $\frac{\partial E}{\partial W}$ by a learning rate ($\eta$). This learning rate determines how quickly the weights will change in each iteration. Because the learning rate is a scalar value, the multiplication operation used is element-wise multiplication.
\\
\[
W = W - (\frac{\partial E}{\partial W} \cdot \eta)
\]
\subsection{Modifying Subsequent Layers}
The modification process for the rest of the layers is similar to the last layer ($L_3$). $\frac{\partial Net}{\partial W}$ and $\frac{\partial \psi}{\partial Net}$ are computed in the exact same way. There is no direct target value, so $\frac{\partial E}{\partial \psi}$ must be computed by taking the dot product of the $\delta$ value from the last layer used in the algorithm (e.g. $L_3$) and the weights on the current layer (e.g. $L_2$).\\
For the first layer of the network, $\frac{\partial Net}{\partial W}$ is  the raw input vector, as there are no weights attached to it. Otherwise, the values are computed using the procedure shown in \ref{procedure}.
\\
\[
\frac{\partial E}{\partial \psi} = \delta_{L-1} \cdot W^T
\]
\\
The weights are modified by repeating this process for the rest of the layers until the original input is reached.
\newpage
\section{Implementation}
Weights do not need to be updated after every iteration during batch training, making it well suited to parallelization. Information is not shared between the calculations for each batch input, so they can each be run in isolation. The input vector for batch training is of dimension (b, n) rather than explicitly (1, n), where n is the input size and b is the number of inputs per batch. Distribution on multiple targets can be implemented by segmenting the input vector into a number of small batches, each of which can be trained in isolation and the resulting $\Delta W$ values can be combined afterwards.
\\
\\
A significant drawback to batch training is that a larger batch size leads to lower network accuracy. Each $\Delta W$ is not influenced by the updates from any other sample in its batch meaning large batch sizes can easily overshoot an optimal error value. The effect is similar to having a learning rate that is too large. These effects can be mitigated through adaptive optimization algorithms such as Adagrad or Adam$^{[3]}$, but ultimately, accuracy will still decrease with large batches.
\\
\\
There is no quantitative way to determine the optimal batch size for any training set. The most common batch sizes range between 16 and 64$^{[4]}$. Smaller batch sizes are more complex to distribute over large systems, leading to a lower efficiency advantage. Large batch sizes tend to degrade accuracy by overadjusting due to the accumulation of $\Delta W$ values.
\newpage
\subsection*{References}
\footnotesize {[1] 1A. Agarwal, S. N. Negahban and M. J. Wainwright, "Stochastic optimization and sparse statistical recovery: An optimal algorithm for high dimensions," 2014 48th Annual Conference on Information Sciences and Systems (CISS), Princeton, NJ, 2014, pp. 1-2.}\\\\
\footnotesize {[2] Pavel Golik, Patrick Doetsch, Hermann Ney, "Cross-Entropy vs. Squared Error Training: a Theoretical and Experimental Comparison", 2013}\\\\
\footnotesize {[3] Sebastian Ruder, "An overview of gradient descent optimisation algorithms", arXiv preprint {\tt arXiv:1609.04747 [cs.LG]}, 2016}\\\\
\footnotesize {[4] Nitish Shirish Keskar, Dheevatsa Mudigere, Jorge Nocedal, Mikhail Smelyanskiy, Ping Tak Peter Tang, "On Large-Batch Training for Deep Learning: Generalization Gap and Sharp Minima", arXiv preprint {\tt arXiv:1609.04836 [cs.LG]}, 2016}\\\\
\footnotesize {Michael A. Nielsen, "Neural Network and Deep Learning", Determination Press, 2015}\\\\
\footnotesize {Matt Mazur, "A Step by Step Backpropagation Example", 2015}\\\\
\footnotesize {Raul Rojas, "Neural Networks", Springer-Verlab, Berlin, 1996}

\vfill

\renewcommand{\abstractname}{Acknowledgements}
\begin{abstract}
\center
This research was supported by AndPlus LLC 
and administered by Abdul Dremali and Vincent Morin.
\end{abstract}

\end{document}